\documentclass{article}

 \usepackage[preprint]{neurips_2026}

\usepackage[utf8]{inputenc} 
\usepackage[T1]{fontenc}    
\usepackage{hyperref}       
\usepackage{url}            
\usepackage{booktabs}       
\usepackage{amsfonts}       
\usepackage{nicefrac}       
\usepackage{microtype}      
\usepackage{xcolor}         
\usepackage{graphicx}
\usepackage{multirow}
\usepackage{amsmath}
\usepackage{wrapfig}
\usepackage{subcaption}
\usepackage[table]{xcolor}
\usepackage[capitalize,noabbrev]{cleveref}
\usepackage{adjustbox}
\newcommand{\rblue}{\rowcolor{blue!10}}
\newcommand{\cblue}{\cellcolor{blue!10}}
\title{Linear-Time Global Visual Modeling \\ \textit{without} Explicit Attention}

\author{
Ruize He$\thanks{Equal Contribution.}$\hspace{4mm}
Dongchen Han$^*$\hspace{2mm}
Gao Huang$\thanks{Corresponding Author.}$
\vspace{3mm}
\\
\small Tsinghua University
}

\begin{document}

\maketitle

\begin{abstract}
    Existing research largely attributes the global sequence modeling capability of Transformers to the explicit computation of attention weights, a process that inherently incurs quadratic computational complexity. In this work, we offer a novel perspective: we demonstrate that attention can be mathematically reframed as a Multi-Layer Perceptron (MLP) equipped with dynamically predicted parameters. Through this lens, we explain attention's global modeling power not as explicit token-wise aggregation, but as an \textit{implicit} process where dynamically generated parameters act as a compressed representation of the global context. Inspired by this insight, we investigate a fundamental question: \textit{can we achieve Transformer-level sequence global modeling entirely through dynamic parameterization while maintaining linear complexity, effectively replacing explicit attention?} To explore this, we design various dynamic parameter prediction strategies and integrate them into standard network layers. Extensive empirical studies on vision models demonstrate that dynamic parameterization can indeed serve as a highly effective, linear-complexity alternative to explicit attention, opening new pathways for efficient sequence modeling. Code is available at \href{https://github.com/LeapLabTHU/WeightFormer}{https://github.com/LeapLabTHU/WeightFormer}.
\end{abstract}

\definecolor{pink}{HTML}{f5a689}
\definecolor{purple}{HTML}{7f71b8}

\section{Introduction}
\begin{wrapfigure}{r}{0.5\textwidth}
    \centering
    \vskip -0.6cm
    \includegraphics[width=0.5\textwidth]{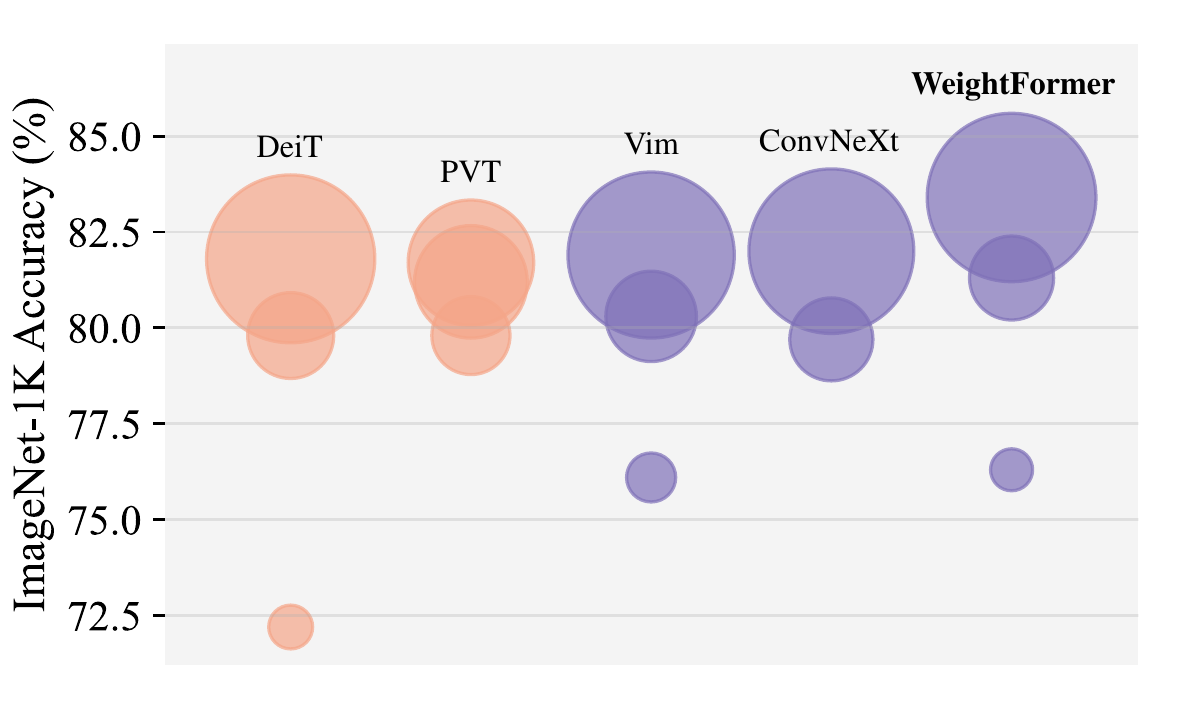}
    \vskip -0.4cm
    \caption{ImageNet accuracy of WeightFormer and baselines. $\color{pink}\circ$ represent models with $O(N^2)$, while $\color{purple}\bullet$ represent models with $O(N)$. Each bubble's area is proportional to FLOPs.}
    \label{fig:flops}
    \vskip -0.5cm
\end{wrapfigure}

The success of Transformers across various domains is primarily rooted in their capacity for global sequence modeling, which is conventionally attributed to the attention mechanism. Under this prevailing perspective, attention is conceptualized as an explicit token-wise weighted aggregation process: attention weights are computed from pairwise token similarities (e.g., $A=\mathrm{Softmax}(QK^\top)$), and then used to explicitly recombine value representations (e.g., $O=AV$). Consequently, efforts to improve efficiency have predominantly focused on approximating or sparsifying this explicit attention matrix~\cite{child2019generating,beltagy2020longformer,zaheer2020big,wang2020linformer,katharopoulos2020transformers,choromanski2020rethinking,han2025vision,yuan2025native,xiong2021nystromformer}, fundamentally bounding the design space to explicit feature recombination.

\begin{wrapfigure}{r}{0.5\textwidth}
    \centering
    \vskip -0.6cm
    \includegraphics[width=0.5\textwidth]{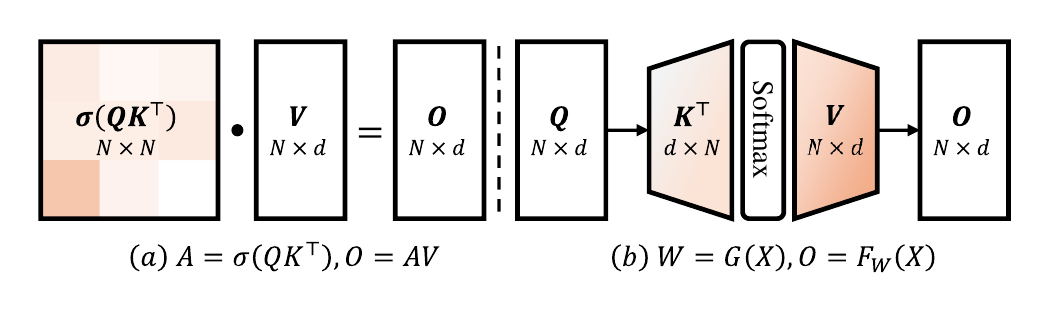}
    \vskip -0.2cm
    \caption{(a) Explicit weighted aggregation, where attention weights are computed and applied to values. (b) Attention as a dynamic MLP with parameters $W=G(X)$, where $K^\top$ and $V$ are dynamic weights and Softmax acts as non-linearity, enabling implicit global modeling via $O=F_W(X)$.}
    \label{fig:dyn}
    \vskip -0.75cm
\end{wrapfigure}

In this work, we challenge this prevailing assumption by introducing a novel perspective: attention can be mathematically reframed as a dynamic MLP. As illustrated in \cref{fig:dyn}, the key matrix $K^\top$ and value matrix $V$ can be viewed as the weights of the first and second linear layer in an MLP, while the Softmax operation serves as its activation function. Crucially, these parameters are not static; they are dynamically generated conditioned on the input.

This perspective shift provides a compelling new explanation for how attention achieves global sequence modeling. Instead of relying on explicit token-to-token routing, global modeling emerges implicitly: the dynamically predicted parameters act as a compact representation that compresses the global context of the input. Forwarding the input through this dynamically parameterized network integrates long-range dependencies naturally, without ever needing to compute or apply explicit attention weights.

This insight leads to our central research question:

\textit{If global modeling is essentially a byproduct of dynamic parameterization, can we use dynamic parameters to completely replace attention, thereby achieving sequence global modeling while maintaining linear complexity?}

To answer this question, we explore whether the implicit global modeling capabilities of attention can be decoupled from its quadratic matrix multiplications. We design and analyze several lightweight mechanisms for dynamically predicting network parameters (e.g., linear and depthwise convolution layers) based on global sequence context. By generating weights dynamically, we bypass the need for an $N \times N$ token interaction matrix, preserving linear computational complexity with respect to the sequence length.

It is important to note that our goal in this paper is not to engineer a new state-of-the-art visual architecture heavily optimized for specific benchmarks. Instead, we aim to systematically explore and validate the feasibility of using dynamic parameters as a fundamental replacement for attention. We instantiate our dynamic strategies within prototype vision models to empirically test this hypothesis.

Extensive experiments demonstrate that models relying solely on our dynamic parameterization strategies can achieve competitive global receptive fields and representational power comparable to Vision Transformers, but with significantly lower computational overhead.

In summary, our contributions are:
\begin{itemize}
    \item We mathematically reframe attention as a dynamic MLP, providing a new explanation for its global sequence modeling capabilities based on implicit parameter generation rather than explicit weight aggregation.
    \item Inspired by this perspective, we explore the feasibility of replacing attention with dynamic parameterization to achieve global modeling while maintaining linear complexity.
    \item We introduce and evaluate various dynamic parameter strategies within vision models. Our results validate that dynamic parameters can effectively substitute attention for global sequence modeling, paving the way for fundamentally more efficient architectures.
\end{itemize}

\section{Related Work}

\paragraph{Attention and Global Modeling Paradigms.}
Attention ~\cite{vaswani2017attention,devlin2019bert,brown2020language,dosovitskiy2020image,touvron2021training} has become the dominant mechanism for global modeling in Transformers, where pairwise token similarities are explicitly computed to produce attention weights that reweight value representations. A substantial line of research has focused on improving the efficiency of attention by modifying the computation of the attention matrix. Representative approaches include sparse attention~\cite{child2019generating,beltagy2020longformer,zaheer2020big,yuan2025native}, low-rank approximations~\cite{wang2020linformer,xiong2021nystromformer}, structured attention patterns~\cite{han2025vision} and kernelized attention~\cite{katharopoulos2020transformers,choromanski2020rethinking}, which aim to reduce the quadratic complexity of the attention matrix. Despite their success, these methods fundamentally remain within the conventional attention paradigm: global modeling is achieved through explicitly computing attention weights and applying them to aggregate values. In other words, the core operation is still similarity-based weighting followed by weighted summation.

In contrast, we propose a fundamentally different perspective: attention can be interpreted as a dynamic parameterized MLP, where global information is compressed into input-conditioned parameters, and global modeling emerges implicitly through forwarding the input through this dynamic network. From this view, explicit attention weight computation is not necessary; instead, dynamic parameter prediction itself serves as the mechanism for global modeling.
\paragraph{Dynamic Networks and Connections to Attention.}
\begin{wraptable}{r}{0.5\textwidth}
    \centering
    \footnotesize
    \renewcommand{\arraystretch}{0.88}
    \vskip -0.45cm
    \caption{Summary of weight prediction strategies.}
    \vskip -0.2cm
    \label{tab:dynamic_summary}
    \begin{tabular}{l|l|l}
        \toprule
        Layer & Strategy  & $\Delta W$                                                              \\
        \midrule
        \multirow{5}*[-2pt]{Linear}
              & GAP       & $\mathrm{MLP}(\mathrm{GAP}(X))$                                         \\
              & Linear    & $W_1 (X^\top X) W_2$                                                    \\
              & Nonlinear & $\sigma\left(W_1 (X^\top X) W_2\right)$                                 \\
              & Deep      & $W_1\sigma\left(W_2 (X^\top X) W_3\right) W_4$                          \\
              & Bilateral & $W_1\sigma(W_2 X^\top)\sigma(X W_3)\, W_4$                              \\
        \midrule
        \multirow{5}*[-2pt]{DWC}
              & GAP       & $\mathrm{MLP}(\mathrm{GAP}(X))$                                         \\
              & Adaptive  & $\mathrm{MLP}(\mathrm{AAP}(X))$                                         \\
              & Amp-Dir   & $s(X)\cdot \dfrac{\mathrm{MLP}(X')}{\|\mathrm{MLP}(X')\|_F + \epsilon}$ \\
              & Conv      & $\mathrm{AAP}(f(X))$                                                    \\
        \bottomrule
    \end{tabular}
    \vskip -0.4cm
\end{wraptable}

Dynamic neural networks, where parameters adapt based on the input, have long been explored as a means to increase model expressiveness~\cite{ha2016hypernetworks,han2022dynamic}. In vision, this includes dynamic filter generation~\cite{jia2016dynamic}, conditional convolutions~\cite{yang2019condconv}, weight modulation~\cite{perez2018film}, and dynamic depthwise operations~\cite{chen2020dynamic}. Several recent works have drawn connections between attention and dynamic convolutions. For instance, \cite{zhou2023interpret} interprets Vision Transformers as ConvNets equipped with dynamic convolutions. \cite{han2021connection} establishes theoretical and empirical links between local attention and dynamic depthwise convolution. Involution~\cite{li2021involution} inverts standard convolution principles by making kernels spatially specific and channel-agnostic, yielding an operation reminiscent of attention. These approaches highlight the dynamic nature of attention weights, effectively treating them as input-dependent kernels that are explicitly generated and applied to aggregate features. However, they still operate within the classic attention paradigm of computing weights and using them for explicit feature recombination.

Our perspective departs from this view: rather than treating attention weights as dynamic kernels, we interpret $K$ and $V$ themselves as the core dynamic parameters of an MLP-like structure. Global information is compressed into these parameters via input-conditioned prediction, enabling implicit global modeling through a simple forward pass.

\section{Attention as Dynamic Parameterized MLP}


\subsection{Revisiting Attention as Explicit Weighting}
We begin by revisiting the standard formulation of attention. Given an input feature matrix $X \in \mathbb{R}^{N \times d}$, the query, key, and value are computed as $Q = X W_Q, K = X W_K, V = X W_V$, where $W_Q, W_K, W_V \in \mathbb{R}^{d \times d}$ are learnable projection matrices. The output is given by
\begin{equation}
    \mathrm{Attention}(Q,K,V) = \mathrm{Softmax}\left(\frac{Q K^\top}{\sqrt{d}}\right)V.
    \label{eq:attn}
\end{equation}
The prevailing perspective in the field interprets \cref{eq:attn} as a weighted combination of value vectors. By defining an attention matrix $A = \mathrm{Softmax}\left(\frac{Q K^\top}{\sqrt{d}}\right) \in \mathbb{R}^{N \times N}$, the operation becomes $O=AV$. Here, $A_{ij}$ explicitly represents the pairwise similarity of token $j$ to token $i$. This view frames global modeling as a two-step process: (1) explicitly computing the dense affinity matrix $A$, and (2) using $A$ to recombine representations from $V$.

Consequently, efforts to reduce the quadratic complexity of Transformers have largely focused on approximating this matrix $A$. Methods such as sparse attention~\cite{child2019generating,zaheer2020big,beltagy2020longformer,yuan2025native}, low-rank approximations~\cite{wang2020linformer,xiong2021nystromformer}, and kernel-based linear attention~\cite{katharopoulos2020transformers,choromanski2020rethinking} all strive to construct or estimate this weighting matrix more efficiently. However, they remain bound to the paradigm that global context must be modeled through explicit feature re-weighting.

\subsection{Attention as Dynamic Parameterized MLP}

We now introduce an alternative perspective that departs from the explicit weighting paradigm. Rather than viewing attention as computing and applying an $N \times N$ attention matrix, we reinterpret it as a \textit{dynamic parameterized MLP} whose parameters are predicted from the input (illustrated in \cref{fig:dyn}).

Consider the output corresponding to a single query vector $q_i \in \mathbb{R}^{d}$ (the $i$-th row of $Q$). From \cref{eq:attn}, we have $o_i = \mathrm{Softmax}\left(\frac{q_i K^\top}{\sqrt{d}}\right) V$.
This expression reveals a structure directly analogous to a two-layer MLP applied to $q_i$. Specifically, $\frac{q_i K^\top}{\sqrt{d}}$ corresponds to a linear transformation with weight matrix $K^\top$, followed by a Softmax non-linearity, and finally a second linear transformation with weight matrix $V$. Crucially, $K^\top$ and $V$ are not static parameters. They are dynamically generated from the input: $K = X W_K, V = X W_V$. Therefore, for each input sequence, the model instantiates a unique MLP whose parameters are conditioned on the entire input. In other words, attention implements a dynamic MLP: $W = G(X), O = F_{W}(X)$, where $G(\cdot)$ predicts the dynamic parameters $W = \{K^\top, V\}$ from $X$, and $F_W(\cdot)$ denotes the forward pass of the resulting MLP.

From this viewpoint, global context is \textit{compressed into the dynamically predicted parameters}. This compression can be interpreted as reducing the representation from the full channel dimension to the individual head dimension, as we conceptualize each head as an independent MLP operating on the global sequence. By forwarding the input through these parameters, the model implicitly integrates global dependencies, without explicitly constructing or applying an attention weight matrix.

\subsection{Implicit Global Modeling via Dynamic Parameters}

This reinterpretation reveals that attention models global context without explicit token routing: global information is implicitly encoded in the dynamically parameters. First, a \textit{global compression} step maps the input $X$ to weights $W = G(X)$, distilling global context. While standard attention compresses channel-wise into subspaces, our approach further employs spatial compression to obtain a fixed-size descriptor, decoupling parameter generation from sequence length. Second, \textit{implicit integration} passes $X$ through the resulting dynamic network $F_W(\cdot)$, naturally fusing long-range dependencies in the forward pass without ever materializing an explicit attention matrix. Notably, this process does not necessitate the explicit computation of attention weights or pairwise token interactions.

\subsection{Implications for Complexity and Model Design}

\textit{Explaining Quadratic Complexity.}
In this formulation, the effective width of the dynamic MLP scales with the number of tokens $N$, since $K^\top \in \mathbb{R}^{d \times N}$ and $V \in \mathbb{R}^{N \times d}$. As a result, the dynamic network grows with sequence length, naturally leading to the quadratic computational and memory complexity of attention. This offers a principled explanation of Transformer scaling behavior.

\textit{Beyond Attention.}
More importantly, this view suggests that explicit attention weights are not essential for global modeling. Instead, global modeling can be achieved through dynamic parameter prediction and implicit computation. This insight opens a new design space for efficient architectures.

\textit{Proving the Feasibility of Replacing Attention with Dynamic Parameters.}
Inspired by this perspective, we aim to validate the feasibility of using dynamic parameterization as a complete replacement for explicit attention in achieving Transformer-level global sequence modeling. To this end, we extend dynamic parameterization to convolutional networks by generating the weights of linear and depthwise convolution layers conditioned on the global input context. This enables standard CNN to implicitly capture global dependencies. The resulting design forms the foundation of WeightFormer, which systematically explores the viability of replacing attention with dynamic parameterization.

\section{Dynamic Weight Prediction}
\label{sec:dwp}

Motivated by the dynamic-MLP interpretation of attention, we now instantiate \textit{implicit global modeling} within convolutional networks. In Transformers, global context is compressed into $K$ and $V$, and global interactions emerge implicitly by forward pass. Our goal is to replicate this mechanism in CNNs: we aim to compress global information into input-conditioned linear and depthwise convolution weights, achieving global modeling through forward passes without explicit token-to-token interactions. However, unlike attention which achieves this compression via head-wise dimensionality reduction, our approach introduces spatial compression paradigms to map the global context into a fixed-size parameter space.

This spatial compression is crucial: by projecting the context into a fixed size, the parameter generation process becomes independent of the input resolution, thereby preserving linear complexity. Formally, given an input feature matrix $X \in \mathbb{R}^{N \times d}$, we construct a compact global representation $\phi(X)$ independent of $N$ and predict dynamic parameters from it. We explore two principled paradigms:
\begin{align}
    \text{Pooling:}      \quad \phi_{\text{pool}}(X) = \mathrm{Pool}(X) \in \mathbb{R}^{M \times d}, \quad
    \text{Correlation:}  \quad \phi_{\text{corr}}(X) = X^\top X \in \mathbb{R}^{d \times d}.
\end{align}
Unlike channel-wise compression in attention $(C \to d)$, our dynamic prediction strategies inherently focus on spatial compression. For instance, the pooling paradigm compresses the sequence into a fixed-size ($M \times d$) descriptor, while the correlation paradigm distills global second-order statistics. Based on these principles, we design dynamic parameter prediction strategies for linear and depthwise convolution layers, summarized in \cref{tab:dynamic_summary}.

\begin{table}[t]
    \centering
    \renewcommand{\arraystretch}{0.88}
    \footnotesize
    \caption{Comparison of different dynamic weight prediction strategies for linear and depthwise convolution layers. The baseline employs static weights. ``Dynamic Linear 1/2'' indicates that dynamic weights are applied to the first and/or second linear layer within the MLP.}
    \label{tab:dwp}
    \begin{tabular}{l|l|ccc|c}
        \toprule
        Method   & Prediction Strategy                                            & \#Param     & FLOPs       & Throughput (img/s) & Acc $\uparrow$ \\
        \midrule
        Baseline & Static CNN                                                     & 5.7M        & 1.1G        & 4172               & 73.3           \\
        \midrule
        \multirow{6}{*}{Dynamic Linear 1}
                 & GAP~\eqref{linear:gap}                                         & 10.1M       & 1.1G        & 3759               & 74.7           \\
                 & Linear w/o Pool~\eqref{linear:linear}                          & 6.8M        & 1.5G        & 2951               & 75.8           \\
                 & Linear w/ Pool~\eqref{linear:linear}                           & 6.8M        & 1.2G        & 3278               & 76.2           \\
                 & Nonlinear~\eqref{linear:nonlinear}                             & 6.8M        & 1.2G        & 3198               & 76.0           \\
                 & Deep~\eqref{linear:deep}                                       & 6.9M        & 1.2G        & 3267               & 76.2           \\
                 & \cblue Bilateral~\eqref{linear:bi}
                 & \cblue 6.9M
                 & \cblue 1.2G
                 & \cblue 3213
                 & \cblue 76.4                                                                                                                      \\
        \midrule
        Dynamic Linear 2
                 & Bilateral~\eqref{linear:bi}                                    & 6.9M        & 1.2G        & 3219               & 75.5           \\
        \midrule
        Dynamic Linear 1 + 2
                 & Bilateral~\eqref{linear:bi}                                    & 8.2M        & 1.3G        & 2659               & 76.7           \\
        \midrule
        \multirow{4}{*}{Dynamic DWC}
                 & GAP~\eqref{conv:gap}                                           & 7.9M        & 1.1G        & 4119               & 74.1           \\
                 & \cblue Adaptive~\eqref{conv:ada}                               & \cblue 6.2M & \cblue 1.1G & \cblue 4051        & \cblue 74.6    \\
                 & \cblue Amp-Dir~\eqref{conv:amp}                                & \cblue 6.4M & \cblue 1.1G & \cblue 4028        & \cblue 74.8    \\
                 & \cblue Conv~\eqref{conv:conv}                                  & \cblue 6.7M & \cblue 1.3G & \cblue 3765        & \cblue 74.8    \\
        \midrule
        Dynamic Linear 1 + DWC
                 & \cblue Bilateral + Adaptive~\eqref{linear:bi}~\eqref{conv:ada} & \cblue 7.4M & \cblue 1.2G & \cblue 3201        & \cblue 76.8    \\
        \bottomrule
    \end{tabular}
    \vskip -0.4cm
\end{table}
\begin{figure}[t]
    \centering
    \includegraphics[width=\linewidth]{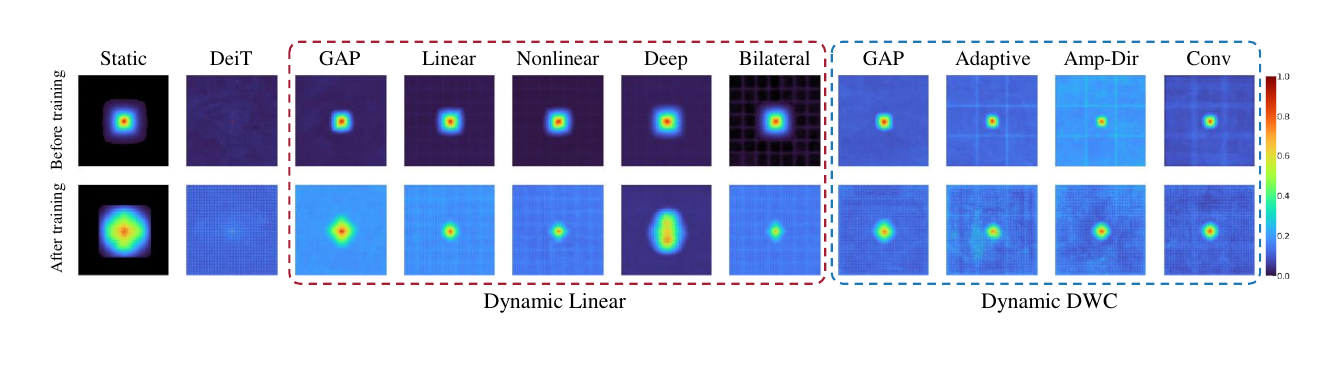}
    \vskip -0.4cm
    \caption{Comparison of ERF~\cite{luo2016understanding} between DeiT, static and dynamic weight prediction strategies. Pixels with higher intensity indicate larger responses related to the central pixel.}
    \label{fig:erf}
    \vskip -0.3cm
\end{figure}
\subsection{Dynamic Linear Layers}
Linear layers operate on channel dimensions and do not perform token-wise mixing. However, when their parameters are conditioned on global input, they can implicitly integrate global context into channel transformations.
For an input $X \in \mathbb{R}^{N \times d}$, we generate a dynamic update $\Delta W(X) \in \mathbb{R}^{d \times d}$ to modulate a learnable static weight $W_0$:
\begin{equation}
    W(X) = W_0 + \Delta W(X).
\end{equation}

\paragraph{Pooling-Based Strategies.}
The simplest approach leverages Global Average Pooling (GAP) to compress the sequence into a vector $z = \mathrm{GAP}(X)$, which is then mapped to parameters via a lightweight MLP:
\begin{equation}
    \Delta W(X) = \mathrm{Reshape}(\mathrm{MLP}(z)).
    \label{linear:gap}
\end{equation}
While efficient, compressing the entire sequence into a single vector risks losing fine-grained features.

\paragraph{Correlation-Based Strategies.}
To capture higher-order feature interactions, we leverage the correlation matrix $X^\top X \in \mathbb{R}^{d \times d}$. Recognizing the limitations of simple linear mappings, we investigate a progression of predictors inspired by the success of deep non-linear architectures:
\begin{align}
    \text{Linear:}    & \quad \Delta W = W_1 (X^\top X) W_2, \label{linear:linear}                          \\
    \text{Nonlinear:} & \quad \Delta W = \sigma\left(W_1 (X^\top X) W_2\right), \label{linear:nonlinear}    \\
    \text{Deep:}      & \quad \Delta W = W_1 \sigma\left(W_2 (X^\top X) W_3\right) W_4. \label{linear:deep}
\end{align}
Specifically, the Nonlinear variant introduces an activation function $\sigma$ (e.g., SiLU). The Deep strategy further factorizes the prediction into a low-rank transition to reduce FLOPs.

Beyond explicitly processing the second-order correlation matrix, we further explore an alternative architectural hypothesis termed \textit{Bilateral Activation}. We attempt to factorize the weight prediction process into two complementary, non-linear branches that act independently on the input $X$ and $X^\top$:
\begin{equation}
    \Delta W(X) = W_1 \sigma(W_2 X^\top) \sigma(X W_3) W_4.
    \label{linear:bi}
\end{equation}
To further reduce computation, we apply Adaptive Average Pooling (AAP) to downsample $X$ by a factor of 2 along the spatial dimensions before computing $X^\top X$. Unless otherwise specified, all experiments adopt this pooling strategy.

\paragraph{Experimental Results.}
\cref{tab:dwp} reports the performance of different linear dynamic weight prediction strategies on ImageNet-1K. Introducing pooling before computing $X^\top X$ not only lowers FLOPs but also slightly improves accuracy, likely by suppressing high-frequency noise while retaining essential global context. Among all strategies, the Bilateral Activation applied to the first linear layer strikes the best accuracy-efficiency trade-off, achieving 76.4\% top-1 accuracy with moderate parameter and FLOPs overhead. Extending it to both linear layers yields further gains (76.7\%), but with higher costs and reduced throughput, suggesting diminishing returns in deeper layers.

\begin{table}[t]
    \centering
    \footnotesize
    \renewcommand{\arraystretch}{0.88}
    \caption{Comparison with Transformer~\cite{touvron2021training,wang2021pyramid,han2024agent}, SSM~\cite{zhu2024vision}, and ConvNet~\cite{he2016deep,liu2022convnet} on ImageNet.}
    \label{tab:class}
    \begin{tabular}{l|c|ccc|c}
        \toprule
        Method                                  & Res.    & \#Params & FLOPs & Throughput$^{224/1024}$ (img/s) & Acc@1 $\uparrow$ \\
        \midrule
        DeiT-T~\cite{touvron2021training}       & 224$^2$ & 6M       & 1.2G  & 3661 / 42                       & 72.2             \\
        DeiT-S~\cite{touvron2021training}       & 224$^2$ & 22M      & 4.6G  & 1469 / 20                       & 79.8             \\
        DeiT-B~\cite{touvron2021training}       & 224$^2$ & 87M      & 17.6G & 446 / 8                         & 81.8             \\
        Agent-DeiT-T~\cite{han2024agent}        & 224$^2$ & 6M       & 1.2G  & 2820 / 160                      & 74.9             \\
        Agent-DeiT-S~\cite{han2024agent}        & 224$^2$ & 23M      & 4.4G  & 1245 / 67                       & 80.5             \\
        PVT-T~\cite{wang2021pyramid}            & 224$^2$ & 13M      & 1.9G  & 1473 / 48                       & 75.1             \\
        PVT-S~\cite{wang2021pyramid}            & 224$^2$ & 25M      & 3.8G  & 1006 / 27                       & 79.8             \\
        PVT-M~\cite{wang2021pyramid}            & 224$^2$ & 44M      & 6.7G  & 672 / 18                        & 81.2             \\
        PVT-L~\cite{wang2021pyramid}            & 224$^2$ & 61M      & 9.8G  & 477 / 13                        & 81.7             \\
        \midrule
        Vim-T~\cite{zhu2024vision}              & 224$^2$ & 7M       & 1.5G  & 1368 / 51                       & 76.1             \\
        Vim-S~\cite{zhu2024vision}              & 224$^2$ & 26M      & 5.1G  & 608 / 23                        & 80.3             \\
        Vim-B~\cite{zhu2024vision}              & 224$^2$ & 98M      & 17.2G & 240 / 9                         & 81.9             \\
        \midrule
        ResNet-50~\cite{he2016deep}             & 224$^2$ & 25M      & 4.1G  & 1639 / 80                       & 76.2             \\
        ResNet-101~\cite{he2016deep}            & 224$^2$ & 45M      & 7.9G  & 978 / 50                        & 77.4             \\
        ResNet-152~\cite{he2016deep}            & 224$^2$ & 60M      & 11.0G & 682 / 36                        & 78.3             \\
        ConvNeXt-S (iso.)~\cite{liu2022convnet} & 224$^2$ & 22M      & 4.3G  & 1517 / 72                       & 79.7             \\
        ConvNeXt-B (iso.)~\cite{liu2022convnet} & 224$^2$ & 87M      & 16.9G & 467 / 23                        & 82.0             \\
        \midrule
        \rblue
        WeightFormer-T                          & 224$^2$ & 7M       & 1.1G  & 3515 / 207                      & 76.3             \\
        \rblue
        WeightFormer-S                          & 224$^2$ & 27M      & 4.4G  & 1226 / 76                       & 81.3             \\
        \rblue
        WeightFormer-B                          & 448$^2$ & 27M      & 17.7G & 1205 / 76                       & 83.4             \\
        \bottomrule
    \end{tabular}
    \vskip -0.5cm
\end{table}

\subsection{Dynamic Depthwise Convolution Layers}

Depthwise convolutions provide spatial inductive bias but are inherently local. By dynamically predicting depthwise kernels from global context, we enable convolutions to achieve Transformer-level global receptive fields. We extend the principle of dynamic parameterization to depthwise convolution layers by predicting input-conditioned depthwise kernels. For an input feature map $X \in \mathbb{R}^{d \times H \times W}$, we generate a kernel update $\Delta W(X) \in \mathbb{R}^{d \times K \times K}$ (where $K=3$) to modulate a learnable static kernel $W_0$:
\begin{equation}
    W(X) = W_0 + \Delta W(X).
\end{equation}
The resulting kernels are applied via depthwise convolution, which naturally maintains linear complexity relative to the spatial resolution. Similar to linear layers, we explore several paradigms to eliminate the dependency on input size.

\paragraph{Global Pooling-Based Strategies.}
A common baseline involves collapsing all spatial dimensions via GAP to obtain a channel descriptor $z = \mathrm{GAP}(X) \in \mathbb{R}^{d}$. An MLP then maps $z$ to the kernel:
\begin{equation}
    \Delta W(X) = \mathrm{Reshape}\left(\mathrm{MLP}(z)\right).
    \label{conv:gap}
\end{equation}
While widely adopted in early dynamic networks~\cite{yang2019condconv, wu2019pay}, this strategy ignores the inherent spatial structure of the input, potentially limiting the adaptivity of the generated filters.

\paragraph{Spatially Adaptive Strategies.}
To preserve structural information, we investigate a \textit{Spatially Adaptive} approach. Instead of a global scalar, we reduce the input to a fixed $K \times K$ resolution that is grid-aligned with the target kernel:
\begin{align}
    X'           = \mathrm{AAP}(X, (K, K)) \in \mathbb{R}^{d \times K \times K}, \quad
    \Delta W(X)  = \mathrm{MLP}(X').
    \label{conv:ada}
\end{align}
Inspired by the stability of weight normalization~\cite{salimans2016weight}, we further explore a \textit{Decoupled Amplitude-Direction (Amp-Dir)} strategy:
\begin{align}
    s(X)         =\mathrm{Sigmoid}(\mathrm{GAP}(X)W), \quad
    \Delta W(X)  = s(X) \cdot \frac{\mathrm{MLP}(X')}{\lVert \mathrm{MLP}(X') \rVert_F + \epsilon}.
    \label{conv:amp}
\end{align}
We also explore predicting depthwise convolution kernels using a convolutional network. The input feature map is first processed by a network composed of two $3\times3$ convolutions with a channel bottleneck and GELU. The resulting feature map is then pooled to the target kernel size:
\begin{equation}
    \Delta W(X) = \mathrm{AAP}(f(X), (K, K)),
    \label{conv:conv}
\end{equation}
where $f(\cdot)$ denotes the convolutional network.

\paragraph{Experimental Results.}
\cref{tab:dwp} reports the performance of different depthwise convolution dynamic weight prediction strategies on ImageNet-1K. Although Amp-Dir and Conv achieve marginal gains (0.2\%), they incur higher parameters, FLOPs, and lower throughput. In contrast, the Spatially Adaptive method preserves spatial structure during kernel prediction while maintaining computational efficiency, offering the best practical trade-off. We therefore adopt it as our depthwise convolution dynamic weight strategy and combine it with Bilateral Activation~\eqref{linear:bi} for the first linear layer, yielding our default design termed ``Dynamic Linear 1 + DWC'' in \cref{tab:dwp}, which achieves the highest accuracy (76.8\%) while remaining efficient.

\paragraph{Validating Global Modeling.}
To verify that our dynamic networks achieve genuine global modeling, we analyze their Effective Receptive Fields (ERF)~\cite{luo2016understanding}. As shown in \cref{fig:erf}, the static baseline remains localized, while all dynamic variants develop expansive receptive fields covering the entire input. This global behavior arises naturally from input-conditioned parameterization. By conditioning weights on global statistics, dynamic weight prediction enables Transformer-like global reasoning while maintaining linear complexity. The evolution from localized responses before training to dense global patterns after training further demonstrates that dynamic weights serve as an efficient alternative for global context aggregation.

\begin{figure}[t]
    \centering
    \includegraphics[width=\linewidth]{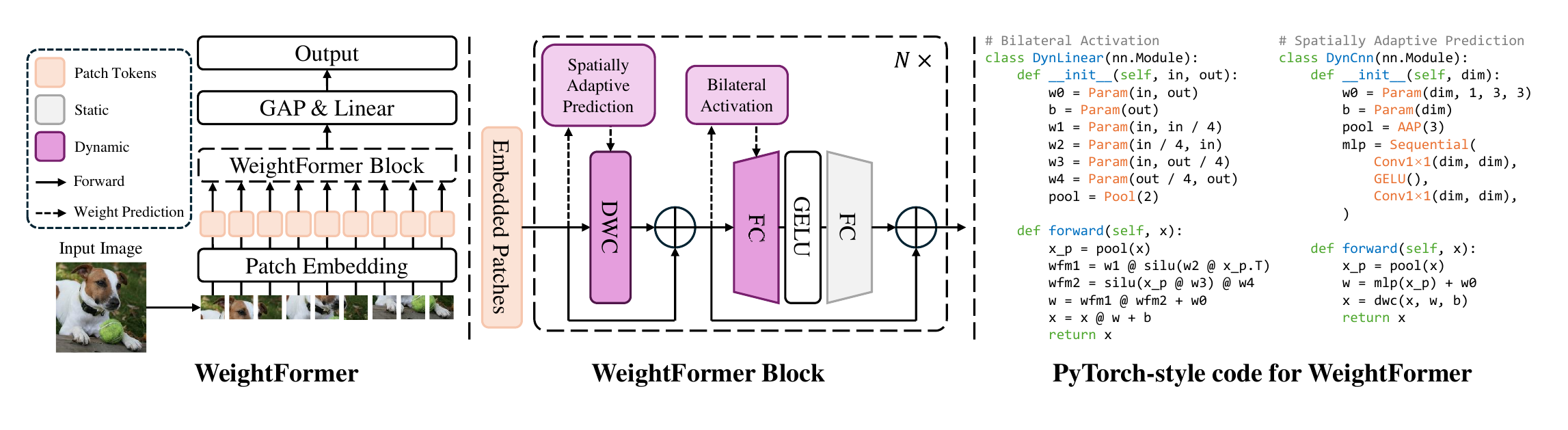}
    \vskip -0.4cm
    \caption{Illustration of WeightFormer architecture. LayerNorm is omitted for simplicity.}
    \label{fig:wfm}
    \vskip -0.6cm
\end{figure}

\begin{table}[t]
    \centering
    \footnotesize
    \renewcommand{\arraystretch}{0.88}
    \caption{Results of object detection and instance segmentation on the COCO val set using Cascade Mask R-CNN~\cite{cai2019cascade} framework. FLOPs$^{t/b}$ denotes total/backbone FLOPs.}
    \label{tab:det}
    \begin{tabular}{l|c|cccccc}
        \toprule
        Backbone                          & FLOPs$^{t/b}$ & AP$^{b}\uparrow$ & AP$^{b}_{50}\uparrow$ & AP$^{b}_{75}\uparrow$ & AP$^{m}\uparrow$ & AP$^{m}_{50}\uparrow$ & AP$^{m}_{75}\uparrow$ \\
        \midrule
        DeiT-T~\cite{touvron2021training} & 594G / 106G   & 44.4             & 63.0                  & 47.8                  & 38.1             & 59.9                  & 40.5                  \\
        \midrule
        \rblue
        WeightFormer-T                    & 566G / 77G    & 45.0             & 63.2                  & 48.4                  & 38.3             & 60.2                  & 40.5                  \\
        \bottomrule
    \end{tabular}
    \vskip -0.5cm
\end{table}

\begin{table}[t]
    \centering
    \footnotesize
    \renewcommand{\arraystretch}{0.88}
    \caption{Results of semantic segmentation. FLOPs$^{t/b}$ denotes total/backbone FLOPs.}
    \label{tab:seg}
    \begin{tabular}{l|l|cc|c}
        \toprule
        Method                             & Backbone                          & \#Params & FLOPs$^{t/b}$ & mIoU $\uparrow$ \\
        \midrule
        DeepLab v3+~\cite{chen2018encoder} & ResNet-101~\cite{he2016deep}      & 63M      & --            & 44.1            \\
        UperNet~\cite{xiao2018unified}     & ResNet-50~\cite{he2016deep}       & 67M      & --            & 41.2            \\
        UperNet~\cite{xiao2018unified}     & ResNet-101~\cite{he2016deep}      & 86M      & --            & 44.9            \\
        \midrule
        UperNet~\cite{xiao2018unified}     & DeiT-T~\cite{touvron2021training} & 11M      & 42G / 11G     & 39.2            \\
        UperNet~\cite{xiao2018unified}     & DeiT-S~\cite{touvron2021training} & 43M      & 157G / 35G    & 44.0            \\
        \midrule
        \rblue
        UperNet~\cite{xiao2018unified}     & WeightFormer-T                    & 12M      & 38G / 7G      & 40.7            \\
        \rblue
        UperNet~\cite{xiao2018unified}     & WeightFormer-S                    & 47M      & 148G / 27G    & 45.6            \\
        \bottomrule
    \end{tabular}
    \vskip -0.5cm
\end{table}

\begin{table}[t]
    \centering
    \renewcommand{\arraystretch}{0.88}
    \footnotesize
    \caption{Ablation on the frequency of dynamic blocks. $N$ denotes the number of dynamic blocks. * refers to the best accuracy before divergence during training.}
    \label{tab:ablation}
    \begin{tabular}{c|ccc|c|l}
        \toprule
        $N$ & \#Params & FLOPs & Throughput (img/s) & Training Loss & Acc   \\
        \midrule
        17  & 10.1M    & 1.3G  & 2323               & --            & 70.2* \\
        14  & 9.3M     & 1.3G  & 2460               & 3.218         & 76.6  \\
        11  & 8.5M     & 1.2G  & 2745               & 3.191         & 76.9  \\
        8   & 7.6M     & 1.2G  & 3218               & 3.253         & 76.2  \\
        \rblue
        6   & 7.1M     & 1.1G  & 3515               & 3.261         & 76.3  \\
        4   & 6.5M     & 1.1G  & 3869               & 3.343         & 75.4  \\
        \bottomrule
    \end{tabular}
    \vskip -0.5cm
\end{table}

\section{WeightFormer: Dynamic Weights For Linear-Time Global Visual Modeling}

Based on the exploration in \cref{sec:dwp}, we introduce \textbf{WeightFormer}, an efficient architecture that selectively incorporates dynamic parameterization to balance modeling capacity and computational efficiency. Instead of uniformly applying dynamic layers, WeightFormer adopts a \textit{sparse distribution strategy}, where dynamic parameterization is inserted every third block, while the rest remain static. This design achieves a favorable trade-off between performance and computational overhead. Applying dynamic parameterization to all blocks would incur substantial cost and hinder fair comparison with the baseline. In contrast, the proposed sparse placement introduces only modest overhead while retaining strong modeling capability. The fixed pattern also ensures a controlled and fair comparison in terms of both parameters and FLOPs. We further validate this design choice in \cref{sec:ablation}.

As illustrated in \cref{fig:wfm}, each dynamic block replaces standard layers with: (1) a dynamic depthwise convolution using Spatially Adaptive Prediction, and (2) an MLP in which only the first linear layer adopts Bilateral Activation. The second linear layer and all static blocks remain unchanged. By decoupling parameter generation from sequence length, WeightFormer maintains strictly linear time and memory complexity, making it well-suited for high-resolution inputs.

\subsection{Image Classification}

The ImageNet-1K~\cite{deng2009imagenet} dataset contains 1.28M training and 50K validation images across 1K classes. Following the Swin Transformer training protocol~\cite{liu2021swin}, all models are trained from scratch for 300 epochs using AdamW~\cite{loshchilov2017decoupled} with cosine learning-rate decay, 20-epoch linear warm-up, weight decay of 0.05, total batch size 2048, and an initial learning rate of $4\times10^{-3}$. Standard augmentations and regularization are applied, including RandAugment~\cite{cubuk2020randaugment}, Mixup~\cite{zhang2018mixup}, CutMix~\cite{yun2019cutmix}, and random erasing~\cite{zhong2020random}. All models were trained on 8 RTX 3090s. \cref{tab:class} compares WeightFormer with strong baselines on ImageNet-1K. Across all model sizes, WeightFormer achieves comparable or superior accuracy while maintaining similar or lower parameter counts and FLOPs. For example, WeightFormer-S achieves 81.3\% accuracy, outperforming DeiT-S (79.8\%) and ConvNeXt-S (iso.) (79.7\%) while using comparable resources. These results demonstrate the feasibility of replacing the explicit attention with dynamic parameters.

\subsection{Object Detection and Instance Segmentation}

We conduct experiments for object detection and instance segmentation on the COCO 2017 dataset~\cite{lin2014microsoft} and use ViTDet~\cite{li2022exploring} as the basic framework. Results are reported in \cref{tab:det}. Compared with DeiT, WeightFormer yields consistent yet modest improvements in both detection and segmentation accuracy, while significantly reducing computational cost. Concretely, WeightFormer-T improves box/mask AP from 44.4/38.1 to 45.0/38.3, while reducing total FLOPs from 594G to 566G and backbone FLOPs from 106G to 77G. These results suggest that dynamic parameterization provides a more efficient way to incorporate global context, leading to improved performance without the heavy computational overhead of attention.

\subsection{Semantic Segmentation}

We evaluate WeightFormer on the ADE20K dataset~\cite{zhou2019semantic} using UperNet~\cite{xiao2018unified} as the segmentation framework. Results are reported in \cref{tab:seg}. WeightFormer-T achieves 40.7 mIoU with 12M params and 38G FLOPs (7G backbone), outperforming DeiT-T (39.2 mIoU) under similar parameter budget yet with markedly lower compute. WeightFormer-S reaches 45.6 mIoU, surpassing DeiT-S by 1.6 points with reduced backbone FLOPs (27G vs. 35G). These gains show that dynamic parameterization strengthens multi-scale semantic modeling efficiently.

\subsection{Image Generation}

\begin{wraptable}{r}{0.5\textwidth}
    \centering
    \renewcommand{\arraystretch}{0.88}
    \footnotesize
    \vskip -0.4cm
    \caption{Results of class-conditional image generation.}
    \vskip -0.2cm
    \label{tab:generation}
    \begin{tabular}{l|cc|c}
        \toprule
        Method                             & \#Params & FLOPs & FID $\downarrow$ \\
        \midrule
        DiT-S/8~\cite{peebles2023scalable} & 33M      & 0.4G  & 153.60           \\
        DiT-S/4~\cite{peebles2023scalable} & 33M      & 1.4G  & 100.41           \\
        DiT-S/2~\cite{peebles2023scalable} & 33M      & 6.1G  & 68.40            \\
        DiT-B/8~\cite{peebles2023scalable} & 131M     & 1.4G  & 122.74           \\
        DiT-B/4~\cite{peebles2023scalable} & 130M     & 5.6G  & 68.38            \\
        DiT-B/2~\cite{peebles2023scalable} & 130M     & 23.0G & 43.47            \\
        \midrule
        DiG-S/2~\cite{zhu2025dig}          & 33M      & 4.3G  & 62.06            \\
        DiG-B/2~\cite{zhu2025dig}          & 131M     & 17.1G & 39.50            \\
        \midrule
        \rblue
        WeightFormer-S/8                   & 38M      & 0.3G  & 149.49           \\
        \rblue
        WeightFormer-S/4                   & 38M      & 1.3G  & 93.05            \\
        \rblue
        WeightFormer-S/2                   & 38M      & 5.0G  & 61.39            \\
        \rblue
        WeightFormer-B/8                   & 150M     & 1.4G  & 122.66           \\
        \rblue
        WeightFormer-B/4                   & 149M     & 5.1G  & 65.89            \\
        \rblue
        WeightFormer-B/2                   & 149M     & 20.0G & 38.21            \\
        \bottomrule
    \end{tabular}
    \vskip -1.2cm
\end{wraptable}

\cref{tab:generation} reports FID results on ImageNet-1K for class-conditional image generation, comparing WeightFormer with DiT~\cite{peebles2023scalable} and DiG~\cite{zhu2025dig}. Across all configurations, WeightFormer consistently reduces FID, indicating improved sample quality. This suggests that implicit global modeling via dynamic parameters benefits both discriminative and generative tasks while preserving computational efficiency.

\subsection{Analysis and Ablation}
\label{sec:ablation}

\paragraph{Efficiency Analysis.}

\begin{wrapfigure}{r}{0.45\columnwidth}
    \centering
    \vskip -0.6cm
    \includegraphics[width=0.45\columnwidth]{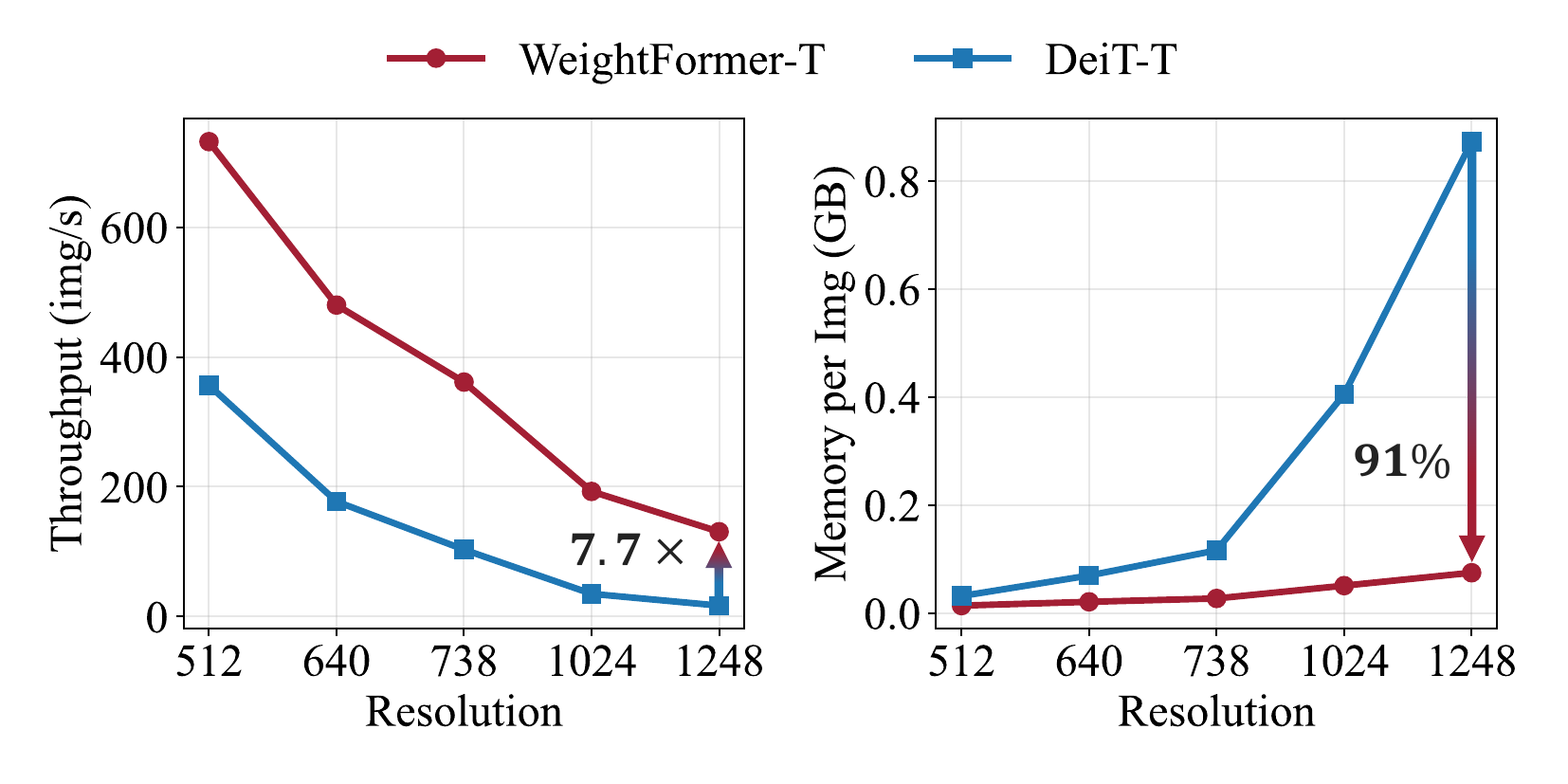}
    \vskip -0.3cm
    \caption{Comparisons between DeiT and WeightFormer in Throughput on RTX~3090, and per-image GPU memory usage.}
    \label{fig:speed}
    \vskip -0.3cm
\end{wrapfigure}

As shown in \cref{tab:class}, WeightFormer achieves higher throughput than strong baselines with competitive accuracy. \cref{fig:speed} further compares throughput and per-image GPU memory. Thanks to its linear complexity, WeightFormer scales well to high resolutions. At 1248$\times$1248 (6,084 tokens), it achieves \textbf{7.7$\times$} higher throughput and \textbf{91\%} memory reduction compared to DeiT, showing strong suitability for high-resolution tasks.

\paragraph{Ablation on Dynamic Block Frequency.}
We study the impact of dynamic block frequency by varying the total number of dynamic blocks, denoted as $N$. As shown in \cref{tab:ablation}, increasing $N$ enhances the model's theoretical capacity but inevitably incurs higher computational costs. Furthermore, replacing too many static layers with dynamic ones does not yield monotonic performance gains; rather, it leads to severe underfitting and optimization challenges. Notably, setting $N=6$ (which corresponds to inserting one dynamic block every third block) strikes the best balance between performance and efficiency.
Consequently, we adopt this sparse distribution strategy as the default configuration for WeightFormer.

\section{Conclusion}
We revisit attention as a form of dynamic MLP with input-conditioned parameters, and explore whether dynamic parameterization can replace explicit attention. Based on this view, we propose WeightFormer. Results on vision tasks show that this approach can achieve competitive performance with improved efficiency. However, this study is still limited in scope. Our evaluation is restricted to vision tasks, and it remains unclear how well this paradigm generalizes to other domains. In addition, the expressivity and inductive biases of dynamic parameterization, are not yet well understood. Furthermore, optimizing these dynamic parameters introduces non-trivial challenges, as their input-conditioned generation can complicate gradient flow and training stability. Future work includes extending this approach to broader settings, improving weight generation mechanisms, and developing a deeper theoretical understanding. We hope this work motivates further exploration of dynamic parameterization as a potential alternative to attention-based architectures.

\section*{Acknowledgement}
This work is supported in part by the National Key R\&D Program of China under Grant 2024YFB4708200, the National Natural Science Foundation of China under Grants U24B20173 and U2541227, and the Scientific Research Innovation Capability Support Project for Young Faculty under Grant ZYGXQNJSKYCXNLZCXM-I20.

We would like to thank Tianyu Li and Zixuan Cao for their valuable preliminary exploration and early-stage investigations that laid a solid foundation for this work.

\bibliographystyle{plain}
\bibliography{main}





\newpage
\appendix
\section{Appendix}

\subsection{Dynamic Weight Strength Analysis}

\begin{figure}[h]
    \centering
    \includegraphics[width=0.6\columnwidth]{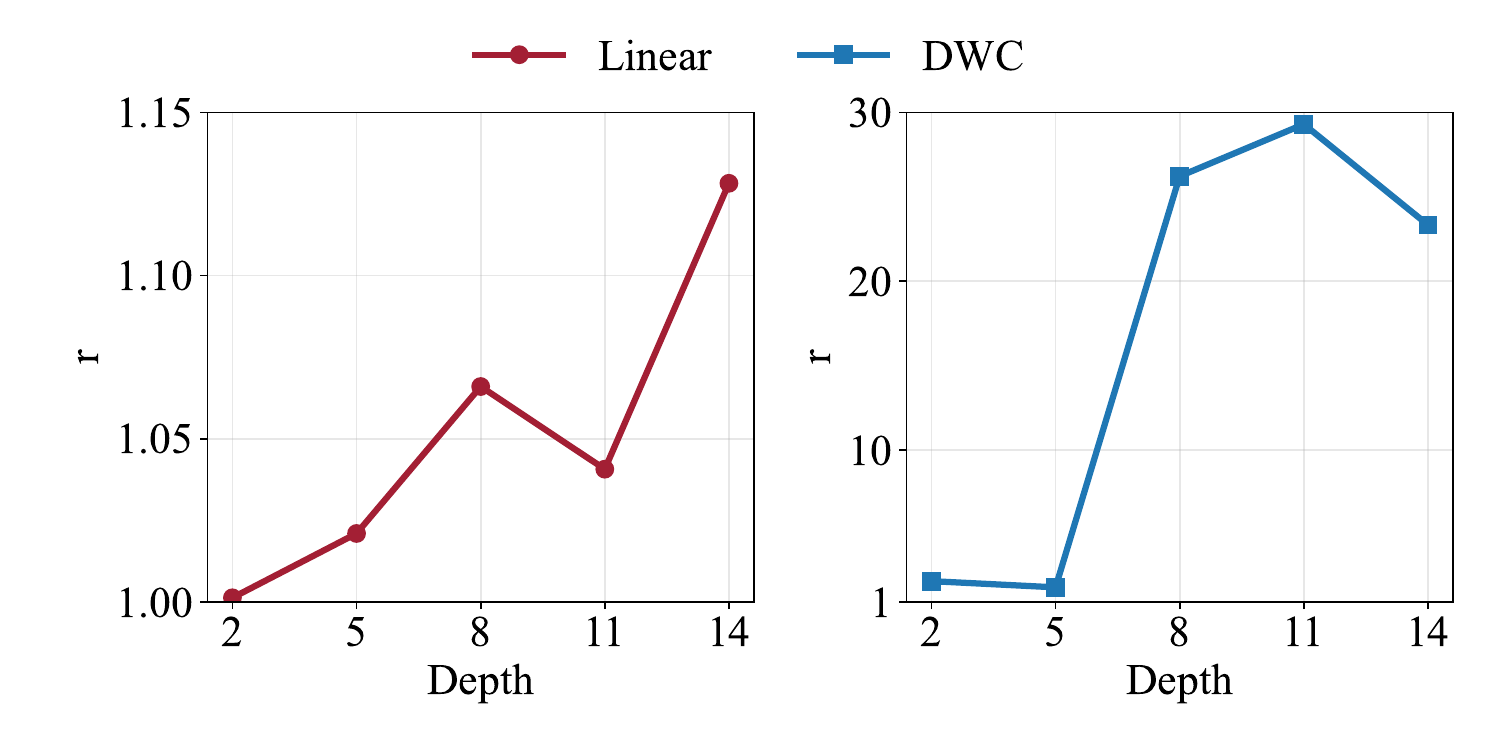}
    \caption{Dynamic weight strength across depth.}
    \label{fig:ratio}
\end{figure}

To analyze how dynamic parameterization contributes across depth, we measure the relative strength of dynamic updates with respect to static parameters. For each dynamic layer, we compute $r = \frac{\lVert \Delta W \rVert_F}{\lVert W_0 \rVert_F}$, where $W_0$ denotes the static weight and $\Delta W$ the predicted dynamic update. This ratio reflects the contribution of input-conditioned parameters. As shown in \cref{fig:ratio}, the relative strength of dynamic parameters exhibits a clear depth-dependent pattern. For dynamic linear layers, the ratio $r$ remains close to 1 across all depths, indicating that channel-mixing transformations are consistently modulated by input-conditioned updates. In contrast, dynamic depthwise convolution exhibits a substantially larger ratio in deeper layers, suggesting increasingly strong spatially adaptive transformations at higher semantic levels. This behavior implies that dynamic depthwise convolutions play a progressively more prominent role in shaping feature representations, while dynamic linear layers provide stable global channel-wise modulation.

\end{document}